\documentclass[letterpaper, 10 pt, conference]{ieeeconf} 

\IEEEoverridecommandlockouts 

\usepackage{cite}

\usepackage{amsmath,amssymb,amsfonts}
\usepackage{graphicx}
\usepackage{textcomp}
\usepackage{xcolor}
\usepackage{color}													
\usepackage{xfrac}	
\usepackage{colortbl}
\usepackage{siunitx}
\usepackage[english]{babel}	
\usepackage{multirow}
\usepackage{hyperref}
\usepackage{blindtext}
\usepackage{mathtools}
\usepackage{hyperref}
\usepackage{tablefootnote}
\usepackage{algorithmic} 
\usepackage{textcomp} 
\usepackage{gensymb} 
\usepackage[ruled, linesnumbered]{algorithm2e}
\usepackage[caption=false, font=footnotesize]{subfig}

\newcommand{\Res}[0]{\boldsymbol{\mathit{\delta}}}
\newcommand{\mm}[1]{\boldsymbol{#1}}
\newcommand{\ind}[1]{\mathrm{#1}}

\title{\LARGE \bf
	Collision Isolation and Identification Using Proprioceptive Sensing\\for Parallel Robots to Enable Human-Robot Collaboration 
}

\author{Aran Mohammad, Moritz Schappler and Tobias Ortmaier
	\thanks{All authors are with the Leibniz University Hannover, Institute of Mechatronic Systems, 30823 Garbsen, Germany,
		{\tt\small aran.mohammad@imes.uni-hannover.de}{\newline Supporting video: \url{https://youtu.be/xD6Zaj6p1f8}}}%
}
\makeatletter
\newcommand{\removelatexerror}{\let\@latex@error\@gobble}
\makeatother

\newif\ifcopyright
		\copyrighttrue
\begin{document}
	
	\ifcopyright
		{\LARGE IEEE Copyright Notice}
		\newline
		\fboxrule=0.4pt \fboxsep=3pt
		
		\fbox{\begin{minipage}{1.1\linewidth}  
				Copyright (c) 2023 IEEE. Personal use of this material is permitted. For any other purposes, permission must be obtained from the IEEE by emailing pubs-permissions@ieee.org. \\
				
				Accepted to be published in: Proceedings of the 2023 IEEE/RSJ International Conference on Intelligent Robots (IROS), October 1 -- 5, 2023, Detroit, Michigan, USA.  
				
		\end{minipage}}
	\else
	\fi

	\graphicspath{{./graphics/}}
	\maketitle
	\thispagestyle{empty}
	\pagestyle{empty}
	
	\begin{abstract}
		Parallel robots (PRs) allow for higher speeds in human-robot collaboration due to their lower moving masses but are more prone to unintended contact. 
		For a safe reaction, knowledge of the location and force of a collision is useful. 
		A novel algorithm for collision isolation and identification with proprioceptive information for a real PR is the scope of this work.
		To classify the collided body, the effects of contact forces at the links and platform of the PR are analyzed using a kinetostatic projection. 
		This insight enables the derivation of features from the line of action of the estimated external force. 
		The significance of these features is confirmed in experiments for various load cases.
		A feedforward neural network (FNN) classifies the collided body based on these physically modeled features. 
		Generalization with the FNN to 300k load cases on the whole robot structure in other joint angle configurations is successfully performed with a collision-body classification accuracy of $84\%$ in the experiments. 
		Platform collisions are isolated and identified with an explicit solution, while a particle filter estimates the location and force of a contact on a kinematic chain. 
		Updating the particle filter with estimated external joint torques leads to an isolation error of less than $\SI{3}{\centi \meter}$ and an identification error of $\SI{4}{\newton}$ in a real-world experiment.
	\end{abstract}
	
	\section{Introduction}
		For a safe human-robot collaboration (HRC), injury levels of unintended contacts are quantified by considering the kinetic energy.
		Design modifications to reduce injuries include lowering the moving masses of lightweight serial robots. Alternatively, parallel robots (PRs) can be used. 
		The drives of a PR are typically fixed to the robot base and are connected to a mobile platform via passive kinematic chains~\cite{Merlet.2006}. 
		Due to lower moving masses, the same energy limits can be maintained at higher speeds.
		\begin{figure}[t!]
			\vspace{1.5mm}
			\centering
			\includegraphics[width=1\columnwidth]{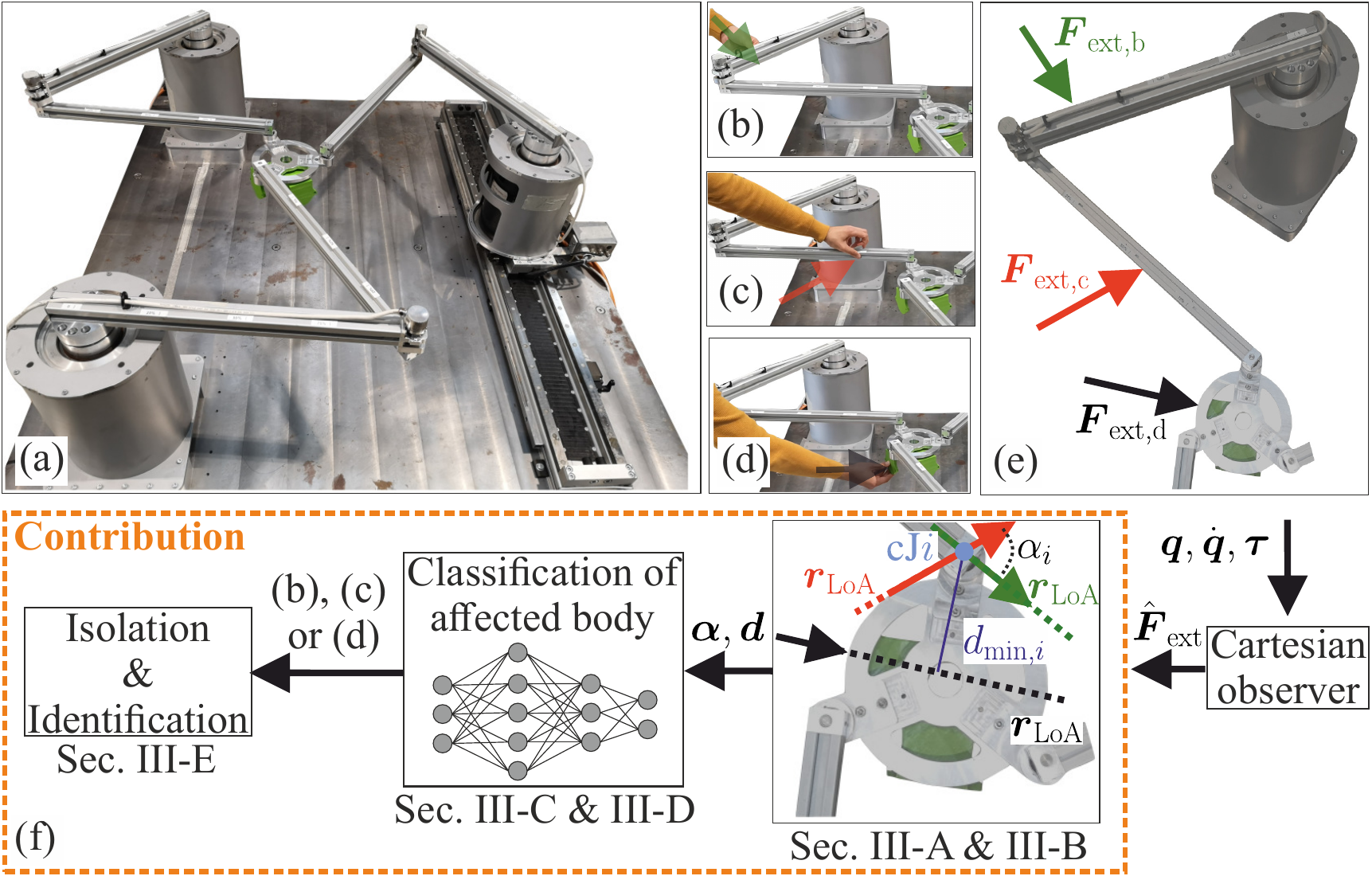}
			\caption{(a) The parallel robot considered in this work with the following collision scenarios: (b) first link, (c) second link, (d) platform, and (e)~the corresponding contact forces. (f) Contribution of this work: based on a Cartesian disturbance observer, the distances $\boldsymbol{d}$ of the line of action $\boldsymbol{r}_\mathrm{LoA}(\lambda)$ to the coupling joints $\mathrm{cJ}$s and the angles $\boldsymbol{\alpha}$ between $\boldsymbol{r}_\mathrm{LoA}(\lambda)$ and the forearms are computed to classify the collided body with an FNN. The second step is the estimation of the location and forces of the collision.}
			\label{fig:titelbild}
			\vspace{-1.5mm}
		\end{figure}
		As an example, the PR considered in this work is shown in Fig.~\ref{fig:titelbild}(a).
		\subsection{Related Work}
			Regardless of the kinematic structure, detection and response to unwanted physical contacts are necessary for HRC.
			Possible collisions between humans and robots are shown in Fig.~\ref{fig:titelbild}(b)--(d). 
			Contact \emph{reactions} for injury reduction require a previous \emph{detection}. 
			This can be done by tactile skin~\cite{Dahiya.2013} or by data-driven modeling for classification into intentional and unintentional contacts~\cite{Golz.2015,Albini.2017,Zhang.2021, Lippi.2021}. 
			Image-based methods allow contact prediction before its occurrence by monitoring the velocity and minimum distance between the human and the robot.
			Preventive reactions incorporate this information into path planning~\cite{Merckaert.2022} or control~\cite{Magrini.2014, Magrini.2015, Luca.2012}. 
			For instance, in~\cite{Hoang.2022}, the contact point between the human and a Hexa PR is determined with a multi-camera system, followed by a recursive Newton-Euler algorithm to calculate the contact force. 
			
			This tactile or visual information must enable detection in dynamic contact scenarios in a fast and robust manner. 
			For this purpose, the use of built-in sensors is more advantageous due to their shorter sample times, lower hardware requirements, and delays.
			In~\cite{Kaneko.1994}, robot movement information is used to locate the contact point as the intersection of the robot configurations in two different iterations. 
			This approach is compared to two other methods on a robot hand in~\cite{Koonjul.2011}. 
			The first considers the joint torques caused by the contact and determines the collided link.
			If compliant control is employed, joint angle displacements allow for data-driven classification of the contact point across the entire robot structure. 
			More powerful machine-learning techniques such as a random forest or feedforward neural networks (FNNs) show the potential to learn the correlation between proprioceptive information and contact position~\cite{Popov.2017, Zwiener.2018}. 
			However, the features must be sampled in a sufficient number of configurations to generalize to unknown contact scenarios.
			 
			The estimation process can also be carried out by optimizing a physically motivated cost function~\cite{Likar.2014, Popov.2019, Wang.2020, Manuelli.2016}. 
			In~\cite{Wang.2020} a velocity-based contact localization is presented. 
			The robot's velocity at the contact point is assumed to only have a tangential component. 
			In~\cite{Manuelli.2016}, multiple contacts on the humanoid robot Atlas are detected and localized by matching the contact position and force to the external joint torques. 
			This is realized by using a contact particle filter, with particles distributed over the entire surface of the robot. 
			As presented in~\cite{Haddadin.2017}, proprioceptive information can also be used to build a physically motivated disturbance observer to detect a contact when the estimated external forces exceed threshold values (\emph{Detection}). 
			Since for serial robots an external force only affects the previous links, the collided link is inferred from the last actuator exceeding the threshold. 
			The observed external moment of the external wrench concerning the link origin is used to determine the line of action (LoA). 
			Its intersection points with the known external hull of the robot are two possible contact locations (\emph{Isolation}). 
			In~\cite{Vorndamme.2021}, an observer based on proprioceptive sensing of the humanoid robot Atlas is analyzed and tested in a simulative study to perform an estimation of contact locations and forces.
		\subsection{Contributions}
			These approaches based on proprioceptive information do not apply to PRs due to their \emph{closed-loop kinematic chains}.
			A contact at one PR chain can excite multiple drives, due to the coupling via the mobile platform.
			Collision isolation and identification for PRs based on proprioceptive information is a research gap addressed in this work.
			As shown in Fig.~\ref{fig:titelbild}(f), a Cartesian disturbance observer allows calculating the minimum distances $\boldsymbol{d}$ of the LoA to the coupling joints, as well as the angles $\boldsymbol{\alpha}$ between the forearms and the LoA. 
			Based on these physically modeled features, an FNN is enabled to classify the collided body in a different configuration. 
			Contact locations and forces on the platform as in Fig.~\ref{fig:titelbild}(d) are determined by an explicit solution, while for the second links (Fig.~\ref{fig:titelbild}(c)) a particle filter is used. 
			In summary, the contributions of this work are:
			\begin{itemize}
				\item Significant features $\boldsymbol{d}, \boldsymbol{\alpha}$ for the collision-body classification of the PR are derived from a kinetostatic analysis.
				\item The hypothesis on the significance of these features for the novel body classification algorithm is experimentally validated using a force-torque sensor and a generalized-momentum observer. 
				\item These features allow classification and generalization to collisions over the entire robot body in unknown configurations. The test of the FNN is performed with 300k data points in other joint angle configurations.
				\item Instead of distributing the particles over the entire PR, the classification result limits the search space of the collision isolation and identification to one body. 
			\end{itemize}
			The structure of the paper is as follows: Section \ref{sec:preliminaries} gives a brief overview of the kinematics and dynamics model for the used PR. 
			The collision isolation and identification algorithms are presented in Sec. \ref{sec:DetAffectedBody}. 
			In Sec. \ref{sec:validation}, the PR used in this work is described, followed by an experimental evaluation of collisions on the whole structure. 
			Finally, this work is concluded in Sec. \ref{sec:conlusions}.
	\section{Preliminaries} \label{sec:preliminaries}
	In this section, the kinematics (\ref{ssec:kinematics}) and dynamics modeling (\ref{ssec:dynamics}), as well as the disturbance observation (\ref{ssec:observer}) are described, summarizing the authors' previous work~\cite{Mohammad.2023}. 
	The modeling is performed exemplarily for the parallel robot used in this work and is generalizable to any fully-parallel robot.
	\begin{figure}[tb!]
		\vspace{1.5mm}
		\centering
		\includegraphics[width=\columnwidth]{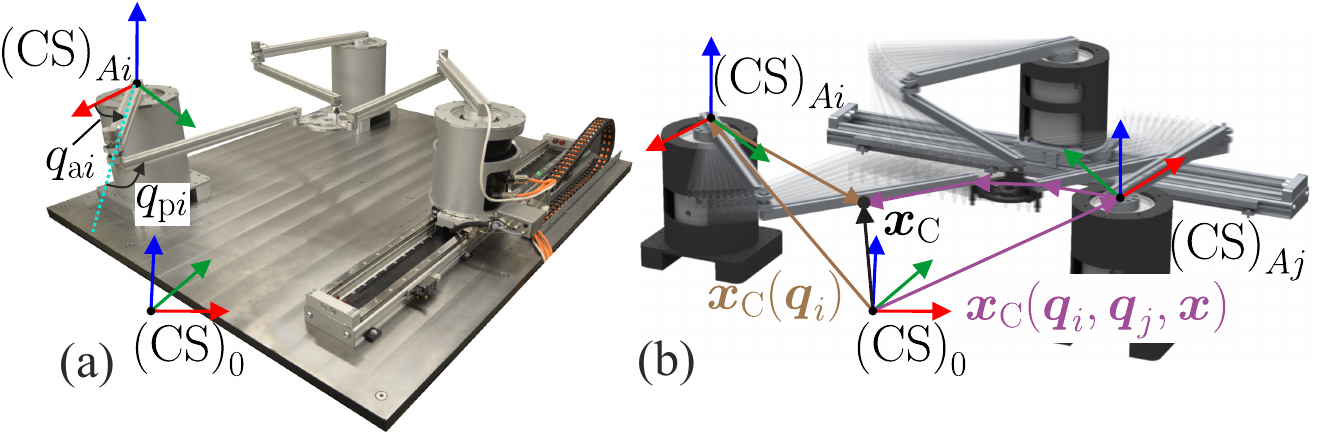}
		\caption{(a) The 3-\underline{R}RR PR from~\cite{Mohammad.2023} (b) with a contact at $\boldsymbol{x}_\mathrm{C}$ at the $i$-th leg chain --- $\boldsymbol{x}_\mathrm{C}$ can be related to any leg chain and the joint angles $\boldsymbol{q}$}
		\label{fig_3PRRR_real_skizze}
		\vspace{-1.5mm}
	\end{figure}
	\subsection{Kinematics} \label{ssec:kinematics}
		The planar 3-\underline{R}RR parallel robot\footnotemark \footnotetext{The letter R denotes a revolute joint and underlining actuation~\cite{Merlet.2006}. The actuated prismatic joint of the PR is kept constant and is therefore not considered in the modeling.}
		shown in Fig.~\ref{fig_3PRRR_real_skizze}(a) consists of $n{=}3$ leg chains with $m{=}3$ platform degrees of freedom~\cite{Thanh.2012}. 
		Operational space coordinates (platform pose), active, passive, and coupling joint angles of the PR are denoted respectively by $\boldsymbol{x}{\in}\mathbb{R}^m,\boldsymbol{q}_\mathrm{a}{\in}\mathbb{R}^n$ and $\boldsymbol{q}_\mathrm{p}, \boldsymbol{q}_\mathrm{c}{\in}\mathbb{R}^3$.
		The $i$-th chains' $n_i{=}3$ joint angles (active, passive, coupling) are represented by $\boldsymbol{q}_i{\in}\mathbb{R}^{n_i}$. 
		A vector containing all $n$ chains' joint coordinates is given via $\boldsymbol{q}^\mathrm{T}{=}[\boldsymbol{q}_1^\mathrm{T}, \boldsymbol{q}_2^\mathrm{T}, \boldsymbol{q}_3^\mathrm{T}] {\in} \mathbb{R}^{3n}$. 
		
		Kinematic constraints $\Res (\boldsymbol{q}, \boldsymbol{x}){=}\boldsymbol{0}$ result from closing vector loops~\cite{Merlet.2006}.
		Reduced kinematic constraints $\Res_\mathrm{red}(\boldsymbol{q}_\mathrm{a}, \boldsymbol{x}){=}\boldsymbol{0}$ and thus active joint angles (inverse kinematics) can be formulated by eliminating the passive joint angles $\boldsymbol{q}_\mathrm{p}$. 
		Passive joint angles are measured to estimate $\boldsymbol{x}$ for the subsequent Newton-Raphson approach (forward kinematics) since the active and passive joint encoder accuracies' vary.
		
		A differentiation w.r.t. time of $\Res{=}\boldsymbol{0}$ and $\Res_\mathrm{red}{=}\boldsymbol{0}$ yields
		\begin{align}\label{eq_DifKin_Jac1}
			\dot{\boldsymbol{q}}&={-}\Res_{\partial \boldsymbol{q}}^{-1}\Res_{\partial \boldsymbol{x}} \dot{\boldsymbol{x}}=\boldsymbol{J}_{q,x}\dot{\boldsymbol{x}}\\
			\label{eq_DifKin_Jac2}
			\dot{\boldsymbol{x}}&= {-}\left(\Res_\mathrm{red}\right)_{\partial \boldsymbol{x}}^{-1} \left(\Res_\mathrm{red}\right)_{\partial \boldsymbol{q}_\mathrm{a}} \dot{\boldsymbol{q}}_\mathrm{a}=\boldsymbol{J}_{x,q_\mathrm{a}}\dot{\boldsymbol{q}}_\mathrm{a}
		\end{align}
		with the Jacobian matrices\footnotemark $\boldsymbol{J}_{q, x}{\in}\mathbb{R}^{3n\times m}$, $\boldsymbol{J}_{x, q_\mathrm{a}}{\in}\mathbb{R}^{m\times n}$ and the notation $\boldsymbol{a}_{\partial \boldsymbol{b}}{\coloneqq} \sfrac{\partial \boldsymbol{a}}{\partial \boldsymbol{b}}$.\footnotetext{For the sake of readability, dependencies on $\boldsymbol{q}$ and $\boldsymbol{x}$ are omitted.}
		
		The kinematics modeling of an arbitrary (contact) point $\mathrm{C}$ with coordinates $\boldsymbol{x}_\mathrm{C}$ on the robot structure is now described using Fig.~\ref{fig_3PRRR_real_skizze}(b).
		Considering the $i$-th kinematic chain, its serial forward kinematics to the point $\mathrm{C}$ equals $\boldsymbol{x}_\mathrm{C}{=}\boldsymbol{f}_i( \boldsymbol{q}_i)$.
		Simultaneously, the formulation $\boldsymbol{x}_\mathrm{C}{=}\boldsymbol{f}_j(\boldsymbol{q}_i, \boldsymbol{q}_j, \boldsymbol{x})$ expresses the contact coordinates by the joint angles $\boldsymbol{q}_j$ of the $j$-th chain and the platform pose $\boldsymbol{x}$, which can be transformed to $\boldsymbol{x}_\mathrm{C}(\boldsymbol{q}_i, \boldsymbol{q}_j)$ by using the rotational constraints in $\Res$~\cite{Schappler.2019}.
		The contact point velocity $\dot{\boldsymbol{x}}_\mathrm{C}{=}\boldsymbol{J}_{x_\mathrm{C},q}\dot{\boldsymbol{q}}$ with the Jacobian matrix $\boldsymbol{J}_{x_\mathrm{C},q}$ is obtained by a time derivative.
		Finally, the differential kinematics between the contact point and respectively the operational space and actuated joint coordinates is derived by using (\ref{eq_DifKin_Jac1}) and (\ref{eq_DifKin_Jac2}) and is formulated via 
		\begin{subequations}\label{eq:contactJacobian} \begin{align}
			\dot{\boldsymbol{x}}_\mathrm{C} &= \boldsymbol{J}_{x_\mathrm{C},q} \dot{ \boldsymbol{q}}\\
			&=\boldsymbol{J}_{x_\mathrm{C},q} \boldsymbol{J}_{q,x} \dot{\boldsymbol{x}} = \boldsymbol{J}_{x_\mathrm{C}, x} \dot{\boldsymbol{x}} \\ 
			&= \boldsymbol{J}_{x_\mathrm{C}, x} \boldsymbol{J}_{x, q_\mathrm{a}} \dot{\boldsymbol{q}}_\mathrm{a} = \boldsymbol{J}_{x_\mathrm{C}, q_\mathrm{a}} \dot{\boldsymbol{q}}_\mathrm{a}
		\end{align} \end{subequations}
		with the Jacobian matrices $\boldsymbol{J}_{x_\mathrm{C}, x}$ and $\boldsymbol{J}_{x_\mathrm{C}, q_\mathrm{a}}$.
	\subsection{Dynamics}\label{ssec:dynamics}
		By noting generalized forces (including moments) in platform coordinates by $\boldsymbol{F}{\in}\mathbb{R}^m$, the equations of motion
		\begin{equation} \label{eq_dyn}
			\boldsymbol{M}_x \ddot{\boldsymbol{x}}{+} \boldsymbol{c}_x {+} \boldsymbol{g}_x{+} \boldsymbol{F}_{\mathrm{fr},x}= \boldsymbol{F}_\mathrm{m} {+} \boldsymbol{F}_\mathrm{ext}
		\end{equation}
		in the operational space and without the constraint forces are obtained by the Lagrangian equations of the second kind, the \textit{subsystem} and \textit{coordinate partitioning} methods~\cite{Thanh.2009}.
		Equation \ref{eq_dyn} contains $\boldsymbol{M}_x$ as the symmetric positive-definite inertia matrix, $\boldsymbol{c}_x{=}\boldsymbol{C}_x\dot{\boldsymbol{x}}$ as the vector/matrix of the centrifugal and Coriolis terms, $\boldsymbol{g}_x$ as the gravitational components, $\boldsymbol{F}_{\mathrm{fr},x}$ as the viscous and Coulomb friction effects, $\boldsymbol{F}_\mathrm{m}$ as the forces based on the motor torques and $\boldsymbol{F}_{\mathrm{ext}}$ as external forces. 
		The projection from forces $\boldsymbol{F}_\mathrm{m}$ into the joint space of the PR is realized by the principle of virtual work $\boldsymbol{\tau}_\mathrm{a}{=}\boldsymbol{J}_{x,q_\mathrm{a}}^\mathrm{T}\boldsymbol{F}_\mathrm{m}$. 
		For a link contact, the projections 
		\begin{subequations}\label{eq_trafo_link_mP_Drives}\begin{align} 
				\boldsymbol{F}_\mathrm{ext,mP}&=\boldsymbol{J}_{x_\mathrm{C},x}^\mathrm{T} \boldsymbol{F}_\mathrm{ext,link},\\
				\boldsymbol{\tau}_\mathrm{a,ext}&=\boldsymbol{J}_{x_\mathrm{C},q_\mathrm{a}}^\mathrm{T} \boldsymbol{F}_\mathrm{ext,link},
		\end{align} \end{subequations}
		of an external force $\boldsymbol{F}_\mathrm{ext,link}$ affect the platform and actuators in a configuration-dependent manner.
	\subsection{Generalized-Momentum Observer} \label{ssec:observer}
		A residual of the generalized momentum $\boldsymbol{p}_x{=}\boldsymbol{M}_x \dot{\boldsymbol{x}}$ is chosen from~\cite{Luca.2003} and formulated in the operational space. 
		The time derivative of the residual leads to $\sfrac{\mathrm{d}}{\mathrm{dt}}\hat{\boldsymbol{F}}_\mathrm{ext} {=} \boldsymbol{K}_{\mathrm{o}} (\dot{\boldsymbol{p}}_x {-} \dot{\hat{\boldsymbol{p}}}_x )$ with $\boldsymbol{K}_\mathrm{o}{=}\mathrm{diag}(k_{\mathrm{o},1},k_{\mathrm{o},2}, \dots ,k_{\mathrm{o},m})$ and $k_{\mathrm{o},i}{>}0$ as the observer gains. 
		Substituting $\hat{\boldsymbol{M}}_x\ddot{\boldsymbol{x}}$ in $\dot{\hat{\boldsymbol{p}}}_x$ with a transformation of (\ref{eq_dyn}) and calculating the time integral of $\dot{\hat{\boldsymbol{F}}}_\mathrm{ext}$, it follows 
		\begin{align} \label{eq:mo}
			\hat{\boldsymbol{F}}_\mathrm{ext} &= \boldsymbol{K}_\mathrm{o} \left( \hat{\boldsymbol{M}}_x \dot{\boldsymbol{x}} {-} \int_{0}^t (\boldsymbol{F}_\mathrm{m} {-} \hat{\boldsymbol{\beta}} {+} \hat{\boldsymbol{F}}_\mathrm{ext}) \mathrm{d}\tilde{t} \right), \text{ with} \\ 
			\nonumber
			\hat{\boldsymbol{\beta}} &= \hat{\boldsymbol{g}}_x {+} \hat{\boldsymbol{F}}_{\mathrm{fr},x} {+}( \hat{\boldsymbol{C}}_x {-}\dot{\hat{\boldsymbol{M}}}_x )\dot{\boldsymbol{x}} = \hat{\boldsymbol{g}}_x {+} \hat{\boldsymbol{F}}_{\mathrm{fr},x}{-}\hat{\boldsymbol{C}}_x^\mathrm{T} \dot{\boldsymbol{x}}
		\end{align}
		and $\dot{\hat{\boldsymbol{M}}}_x{=}\hat{\boldsymbol{C}}_x^\mathrm{T}{+}\hat{\boldsymbol{C}}_x$~\cite{Haddadin.2017, Ott.2008} for the generalized-momentum observer (MO). 
		Assuming $\hat{\boldsymbol{\beta}}{\approx}\boldsymbol{\beta}$, a linear and decoupled error dynamics $\boldsymbol{K}_\mathrm{o}^{-1} \dot{\hat{\boldsymbol{F}}}_\mathrm{ext} {+} \hat{\boldsymbol{F}}_\mathrm{ext}{=}\boldsymbol{F}_\mathrm{ext}$ in the operational space applies.

	\section{Isolation and Identification} \label{sec:DetAffectedBody}
		Effects of collisions on the mobile platform (\ref{ssec:Coll_mP}) and the links of a kinematic chain (\ref{ssec:Coll_Link}) are presented at the beginning of this section. 
		The classification algorithm is described afterward with a decision tree (\ref{ssec:ClassAlg_DT}) and an FNN~(\ref{ssec:ClassAlg_FNN}).
		Finally, the particle filter is introduced (\ref{ssec:IsolItend_PF}).
		\subsection{Collision on the Mobile Platform} \label{ssec:Coll_mP} 
			A contact wrench $\boldsymbol{F}_\mathrm{ext}{=}(\boldsymbol{f}^\mathrm{T}, \boldsymbol{m}^\mathrm{T})^\mathrm{T}$ consisting of forces $\boldsymbol{f}$ and moments $\boldsymbol{m}$ at the mobile platform is considered, and $\hat{\boldsymbol{F}}_\mathrm{ext}, \hat{\boldsymbol{\tau}}_{\mathrm{a,ext}}$ are estimated by the MO. 
			\begin{figure}[t!]
				\vspace{1.5mm}
				\centering
				\includegraphics[width=\columnwidth]{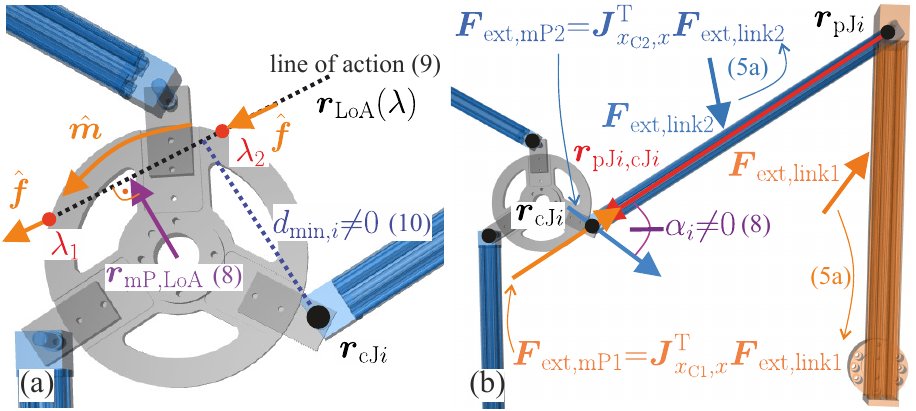}
				\caption{(a) Effects of an external force $\boldsymbol{f}$ on the platform with the estimate $\hat{\boldsymbol{F}}_\mathrm{ext}{=}(\hat{\boldsymbol{f}}^\mathrm{T}, \hat{\boldsymbol{m}}^\mathrm{T})^\mathrm{T}$, the minimum lever $\boldsymbol{r}_\mathrm{mP,LoA}$, the line of action $\boldsymbol{r}_\mathrm{LoA}(\lambda)$ and intersection points at $\lambda_1,\lambda_2$ in a MuJoCo simulation~\cite{Todorov.2012}. 
				The minimum distance $d_{\min,i}$ is between $\boldsymbol{r}_\mathrm{LoA}(\lambda)$ and the coupling point $\boldsymbol{r}_{\mathrm{cJ}i}$. 
				(b) Link forces $\boldsymbol{F}_{\mathrm{ext,link}}$ with their projections $\boldsymbol{F}_{\mathrm{ext,mP}}$ on the platform coordinates. The angle $\alpha_i$ is included by the projection $\boldsymbol{F}_\mathrm{ext,mP2}$ and the vector $\boldsymbol{r}_{\mathrm{pJ}i,\mathrm{cJ}i}$}
				\label{fig:KinetostaticAnalysis}
				\vspace{-1.5mm}
			\end{figure}
			In Fig.~\ref{fig:KinetostaticAnalysis}(a) the procedure of the following section is shown. 
			The first assumption is made with $\boldsymbol{m}{=}\boldsymbol{0}$ so that the equation
			\begin{equation} 
				\begin{split}
					\hat{\boldsymbol{m}} = \boldsymbol{r} {\times} \hat{\boldsymbol{f}} = \boldsymbol{S} 	(\boldsymbol{r})\hat{\boldsymbol{f}} = \boldsymbol{S}^\mathrm{T} ( \hat{\boldsymbol{f}}) \boldsymbol{r}
				\end{split}
			\end{equation}
			holds with $\boldsymbol{S}$ as a skew-symmetric matrix operator and $\boldsymbol{r}$ as a lever between the body-fixed platform coordinate system to any point on the LoA $\boldsymbol{r}_\mathrm{LoA}(\lambda)$. 
			Using the Moore-Penrose inverse~$(\dagger)$ of $\boldsymbol{S}(\hat{\boldsymbol{f}})$~\cite{Haddadin.2017}, the minimum distance 
			\begin{equation} \label{eq:CalcMPLoA}
				\begin{split} 
					\boldsymbol{r}_\mathrm{mP,LoA} &= ( \boldsymbol{S}^\mathrm{T} ( \hat{\boldsymbol{f}}) )^\dagger \hat{\boldsymbol{m}}
				\end{split}
			\end{equation} 
			from the platform coordinate system to $\boldsymbol{r}_\mathrm{LoA}(\lambda)$ of the external force is calculated. 
			Now the LoA 
			\begin{equation} \label{eq:CalcLoA}
				\boldsymbol{r}_\mathrm{LoA}(\lambda){=}\boldsymbol{r}_\mathrm{mP, LoA}{+}\lambda \hat{\boldsymbol{n}}_\mathrm{f}
			\end{equation} with $\hat{\boldsymbol{n}}_\mathrm{f}{=}\hat{\boldsymbol{f}}/||\hat{\boldsymbol{f}} ||_2$ and the scalar 	variable $\lambda$ can be determined, leading to the two intersections $\lambda_1, \lambda_2$ with the known platform hull. 
			These two cases correspond to a pull ($\lambda_1$) and push ($\lambda_2$) force. 
			Unwanted contacts are assumed to be the latter. 			
			Thus the contact location at the mobile platform is determined and together with the estimation of the MO in (\ref{eq:mo}) the collision isolation and identification for platform contacts are completed.
		\subsection{Collision at a Link} \label{ssec:Coll_Link}
			Figure~\ref{fig:KinetostaticAnalysis}(b) depicts the two link forces $\boldsymbol{F}_\mathrm{ext,link1}$, $\boldsymbol{F}_\mathrm{ext,link2}$ at the first and second link of chain $i$ and their projections $\boldsymbol{F}_\mathrm{ext,mP1}$, $\boldsymbol{F}_\mathrm{ext,mP2}$ to the platform coordinates with the corresponding Jacobian matrix $\boldsymbol{J}_{x_{\mathrm{C}},x}$ from (\ref{eq_trafo_link_mP_Drives}).		
			The link collision in Fig.~\ref{fig:KinetostaticAnalysis}(b) differs from the platform contact in Fig.~\ref{fig:KinetostaticAnalysis}(a) since the minimum distance 
			\begin{equation} \label{eq:MinDistd}
				\begin{split} 
				d_{\mathrm{min},i}{=}||\left(\boldsymbol{r}_{\mathrm{cJ}i}{-}\boldsymbol{r}_\mathrm{mP, LoA}\right) {\times} \hat{\boldsymbol{n}}_\mathrm{f}||_2
				\end{split}
			\end{equation} 
			from $\boldsymbol{r}_\mathrm{LoA}(\lambda)$ to the $i$-th coupling joint $\boldsymbol{r}_{\mathrm{cJ}i}$ is zero.
			This allows the determination of the leg chain on which the force acts. 
			Since the link contact force affects the platform via the passive revolute coupling joints of the $i$-th kinematic chain, the force's projection in platform coordinates intersects with the coupling joint.
			
			The vector $\boldsymbol{r}_{\mathrm{pJ}i,\mathrm{cJ}i}$ from the passive joint $\mathrm{pJ}i$ to the coupling joint $\mathrm{cJ}i$ of the $i$-th chain and $\hat{\boldsymbol{n}}_\mathrm{f}$ define the angle
			\begin{equation} \label{eq:Anglealpha}
				\begin{split} 
				\alpha_i{=}\angle (\hat{\boldsymbol{n}}_\mathrm{f}, \boldsymbol{r}_{\mathrm{pJ}i,\mathrm{cJ}i}).
				\end{split}
			\end{equation} 
			The lines of action of $\boldsymbol{F}_\mathrm{ext,mP1},\boldsymbol{F}_\mathrm{ext,mP2}$ show in comparison the difference that $\boldsymbol{F}_\mathrm{ext, mP1}$ with $\alpha_i{=\SI{180}{^\circ}}$ is antiparallel (or with $\alpha_i{=\SI{0}{^\circ}}$ parallel) to $\boldsymbol{r}_{\mathrm{pJ}i,\mathrm{cJ}i}$. 
			In contrast, $\boldsymbol{F}_\mathrm{ext,mP2}$ with $\boldsymbol{r}_{\mathrm{pJ}i,\mathrm{cJ}i}$ includes the angle $\SI{0}{^\circ}{<}|\alpha_i|{<}\SI{180}{^\circ}$. 
			Furthermore, the distinction of links can be made based on $\hat{\boldsymbol{\tau}}_{\mathrm{a,ext}}{=}\boldsymbol{J}_{x,q_\mathrm{a}}^\mathrm{T}\hat{\boldsymbol{F}}_\mathrm{ext}$, since $\boldsymbol{F}_\mathrm{ext,link1}$ only acts on the affected actuated joint\footnote{Isolation for collisions at the first link can only be realized up to the contact body classification since the exact contact location cannot be determined. The difference is that only one drive is excited.}.
			
			\begin{figure}[tb!]
				\removelatexerror
				\vspace{1.5mm}
				\begin{algorithm}[H]
					\caption{Calculate features $\boldsymbol{d}, \boldsymbol{\alpha}, n_\tau $}\label{alg:classifyContBody_geomFeat}
					{\small 
						\SetKwInOut{Input}{Input}
						\SetKwInOut{Output}{Output}
						\Input{$\hat{\mm{F}}_\ind{ext},\hat{\mm{\tau}}_\ind{a,ext}, \mm{q},\mm{x}, \varepsilon_{\hat{\tau}_{\mathrm{a}}}$}
						\Output{collision-body-relevant features $\boldsymbol{d}, \boldsymbol{\alpha}, n_\tau$}
						$\boldsymbol{r}_\mathrm{mP,LoA}\gets$ Minimal lever by (\ref{eq:CalcMPLoA})\;					
						$\boldsymbol{r}_\mathrm{LoA}(\lambda)\gets$ Line of action by (\ref{eq:CalcLoA})\;	
						$n_\tau \gets 0$ Declare variable as the number of affected drives\;
						$\boldsymbol{d} {\gets}\boldsymbol{0}$ Declare array for the minimal distances for $n$ chains\;
						$\boldsymbol{\alpha} {\gets} \boldsymbol{0}$ Declare array for the angles for $n$ chains\;
						\For{$i{=}1$ to $n$} 
						{\tcp{Calculate features}
							$\boldsymbol{r}_{\mathrm{cJ}i}\gets$ $i$-th coupling joint position by serial forward kinematics\;		
							$d_i \gets$ Minimal distance $d_{\min,i}$ by (\ref{eq:MinDistd}) in row $i$ of $\boldsymbol{d}$\;
							$\boldsymbol{r}_{\mathrm{pJ}i,\mathrm{cJ}i}\gets$ Vector from passive joint to coupling joint\;	
							$\alpha_i\gets$ Angle $\alpha_i$ by (\ref{eq:Anglealpha}) in row $i$ of $\boldsymbol{\alpha}$\;
							\If{$|\hat{\tau}_{\mathrm{a}_i\mathrm{,ext}}|{>}\varepsilon_{\hat{\tau}_{\mathrm{a}}}$}
							{
								{$n_\tau\gets n_\tau{+}1$\;}
							}
						}
					}
				\end{algorithm}
				\vspace{-0.5mm}
			\end{figure}
			By these considerations and by generalizing from the special case of Fig.~\ref{fig:KinetostaticAnalysis}, we set up the hypothesis that the model-based estimates $\boldsymbol{d}^\mathrm{T}{=}[d_{\min,1}, \dots, d_{\min,n}], \boldsymbol{\alpha}^\mathrm{T}{=}[\alpha_1, \dots, \alpha_n]$ and $\hat{\boldsymbol{\tau}}_\mathrm{a,ext}$ can be used to classify the collided body of a PR. 
			The robot configuration is implicitly captured in the features $\boldsymbol{d}, \boldsymbol{\alpha}$ and $\hat{\boldsymbol{\tau}}_\mathrm{a,ext}$, ensuring generalization for contacts in new configurations.
			This \emph{reduces the necessity for an extensive sampling} of the high-dimensional configuration space. 
			Algorithm~\ref{alg:classifyContBody_geomFeat} summarizes the calculation of $\boldsymbol{d},\boldsymbol{\alpha}$ and the number $n_\tau$ of affected drives for collision-body classification.
			The inputs $\mm{q}$ and $\mm{x}$ to Alg. \ref{alg:classifyContBody_geomFeat} follow from the kinematics modeling, while $\hat{\boldsymbol{F}}_\mathrm{ext}$ and $\hat{\boldsymbol{\tau}}_{\mathrm{a,ext}}$ are obtained by~(\ref{eq:mo}). 
			\begin{figure}[tb!]
				\removelatexerror
				\vspace{1.5mm}
				\begin{algorithm}[H]
					\caption{Classification with a decision tree}\label{alg:classifyContBody_DT}
					{\small 
						\SetKwInOut{Input}{Input}
						\SetKwInOut{Output}{Output}
						\Input{$\boldsymbol{d}, \boldsymbol{\alpha}, n_\tau, \varepsilon_{d}, \varepsilon_{\alpha}$}
						\Output{predicted body $\hat{k}_\mathrm{body}$}
						\uIf{$\min (\boldsymbol{d}) {<} \varepsilon_{d}$}
						{
							$j \gets$ Index of $\min(\boldsymbol{d})$ in $\boldsymbol{d}$\;			
							$\alpha_j \gets$ $j$-th entry in $\boldsymbol{\alpha}$\;			
							\uIf{$((|\alpha_j|{<}\varepsilon_{\alpha}) \lor 
								(180\degree {-} |\alpha_j|{<}\varepsilon_{\alpha})
								) \land (n_\tau{<}2)$}
							{$\hat{k}_\mathrm{body}\gets$ First link of the $j$-th chain;}
							\Else {$\hat{k}_\mathrm{body}\gets$ Second link of the $j$-th chain;}
						}
						\Else {$\hat{k}_\mathrm{body}\gets$ mobile platform\;}
					}
				\end{algorithm}
				\vspace{-0.5mm}
			\end{figure}
		\subsection{Collision-Body Classification with a Decision Tree} \label{ssec:ClassAlg_DT}		
		Algorithm \ref{alg:classifyContBody_DT} shows the workflow of a threshold method in the form of a simple decision tree (DT) for classifying the collided body $\hat{k}_\mathrm{body}$.
		To handle the influence of modeling inaccuracies, thresholds for the conditions on $\boldsymbol{d},\boldsymbol{\alpha},\hat{\boldsymbol{\tau}}_\mathrm{a,ext}$ in lines 1 and 4 are selected as $\varepsilon_{d}, \varepsilon_{\alpha}$. 
		Line 1 distinguishes a contact between the platform and the respective leg chain by comparing $\min(\boldsymbol{d})$ with $\varepsilon_{d}$. 
		By $\alpha_j, n_\tau$ in line 4, a contact at the first or second link of the $j$-th leg chain can subsequently be identified.
		\subsection{Collision-Body Classification with a Neural Network} \label{ssec:ClassAlg_FNN}
		The presented DT has three parameters $\varepsilon_{d}, \varepsilon_{\alpha},\varepsilon_{\hat{\tau}_{\mathrm{a}}}$. 
		The limited number may result in ambiguous and misclassified cases due to modeling inaccuracies. 
		An FNN is therefore selected as another classification algorithm. 
		The gradient-based optimization method Adam~\cite{Kingma.22122014, FabianPedregosa.2011} is performed to train the FNN with the physically modeled inputs $\hat{\boldsymbol{F}}_\mathrm{ext}$, $\hat{\boldsymbol{\tau}}_\mathrm{a,ext}$, $\boldsymbol{d}$, $\boldsymbol{\alpha}$ and the known contact body $k_\mathrm{body}$ as output. 
		The hyperbolic tangent function is selected as the activation function in the hidden layers. 
		Since the inputs are available in robot operation, real-time prediction is possible.
		An $L_2$ regularization term $\lambda {\ge} 0$ and the network structure with the number of hidden layers and neurons are determined by a grid search in a hyperparameter optimization to avoid underfitting and overfitting. 
		\subsection{Particle Filter for the Second Links} \label{ssec:IsolItend_PF} 
		If a second link is classified as $\hat{k}_\mathrm{body}$, a particle filter with $R$ particles will be initiated for that body only. 
		The $r$-th particle is represented at the $k$-th time step by the vector 
		\begin{align}
			\boldsymbol{p}_k^{[r]}{=}[l_\mathrm{C}^{[r]},\hat{f}_\mathrm{C}^{[r]}]^\mathrm{T}
		\end{align}
		with the estimated contact force $\hat{f}_\mathrm{C}^{[r]}$ at the second link.
		It is assumed that only a contact force orthogonal to the second link occurs in the case of a collision. 
		Here $0 {\le} l_\mathrm{C}^{[r]} {\le} 1$ is a variable normalized to the link length. It is expressed and invariant in the body-fixed joint coordinate system.
		At the passive joint $\mathrm{pJ}i$, $l_\mathrm{C}^{[r]}{=}0$ and increases along the second link to the coupling joint $\mathrm{cJ}i$.
		This allows one-dimensional collision isolation for the planar PR, which is supported by the link length ratio of ${\approx}24$ of total length to the radius.
		In the motion model 
		\begin{align} \label{eq:pf_motionmodel}
			\boldsymbol{p}_{k+1}^{[r]}{\sim}\mathcal{N}(\boldsymbol{p}_{k}^{[r]}, \boldsymbol{\Sigma}_\mathrm{mot}),
		\end{align}
		each particle position is updated by sampling a normal distribution with a covariance matrix $\boldsymbol{\Sigma}_\mathrm{mot}$.
		The measurement model with the importance weights 
		\begin{align} \label{eq:pf_measurementmodel}
			w_k^{[r]}{=}\exp \left(
			{-}\frac{1}{2} 
			( \hat{\boldsymbol{\tau}}_{\mathrm{a,ext}} {-} \hat{\boldsymbol{\tau}}_{\mathrm{a,ext}}^{[r]})^\mathrm{T}
			\boldsymbol{\Sigma}_\mathrm{meas}^{-1} 
			( \hat{\boldsymbol{\tau}}_{\mathrm{a,ext}}{-}\hat{\boldsymbol{\tau}}_{\mathrm{a,ext}}^{[r]} )
		\right)
		\end{align}
		includes a covariance matrix $\boldsymbol{\Sigma}_\mathrm{meas}$ and the estimated external joint torques $\hat{\boldsymbol{\tau}}_\mathrm{a,ext}$.
		In (\ref{eq:pf_measurementmodel}), $\hat{\boldsymbol{\tau}}_{\mathrm{a,ext}}^{[r]}$ represents the projection 
		\begin{align}
			\hat{\boldsymbol{\tau}}_\mathrm{a,ext}^{[r]} = \boldsymbol{J}_{\boldsymbol{x}_\mathrm{C}(l_\mathrm{C}^{[r]}),q_\mathrm{a}}^\mathrm{T} \hat{\boldsymbol{F}}_\mathrm{C}^{[r]}
		\end{align}
		by (\ref{eq_trafo_link_mP_Drives}) of the estimated forces $\hat{\boldsymbol{F}}_\mathrm{C}^{[r]}{=}^0\boldsymbol{R}_{\mathrm{pJ}i} [0,\hat{f}_\mathrm{C}^{[r]},0]^\mathrm{T}$ in the particle with the location $l_\mathrm{C}^{[r]}$ expressed in $\left(\mathrm{CS}\right)_{\mathrm{pJ}i}$ of the $i$-th passive joint onto the actuated joint coordinates.
		Thus, the particle positions are weighted with $w_k^{[r]}$ according to their fit to the estimated external joint torques.
		Finally, an importance resampling is performed according to $w_k^{[r]}$.
	\section{Validation} \label{sec:validation}
	Starting with the description of the experimental setup (\ref{ssec:ExpSetup}), the generalization of the classification algorithm is evaluated in a simulation (\ref{ssec:SimRes}). 
	The isolation and identification of collisions are finally validated experimentally (\ref{ssec:ExpRes}). 
	\subsection{Experimental Setup} \label{ssec:ExpSetup}
		A force-torque sensor\footnote{KMS40 from Weiss Robotics} (FTS) measures the contact force for validation of the proprioceptive collision identification.
		Through a ROS package\footnote{\url{https://github.com/ipa320/weiss_kms40}} of the FTS, the measurement is synchronized in time with the robot control in \textsc{Matlab}/Simulink. 
		The communication is based on the EtherCAT protocol and the open-source tool EtherLab\footnote{\url{https://www.etherlab.org}} with an external-mode patch and a shared-memory real-time interface\footnote{\url{https://github.com/SchapplM/etherlab-examples}}.
		\begin{figure}[b!]
			\vspace{1.5mm} 
			\centering
			\includegraphics[width=\columnwidth]{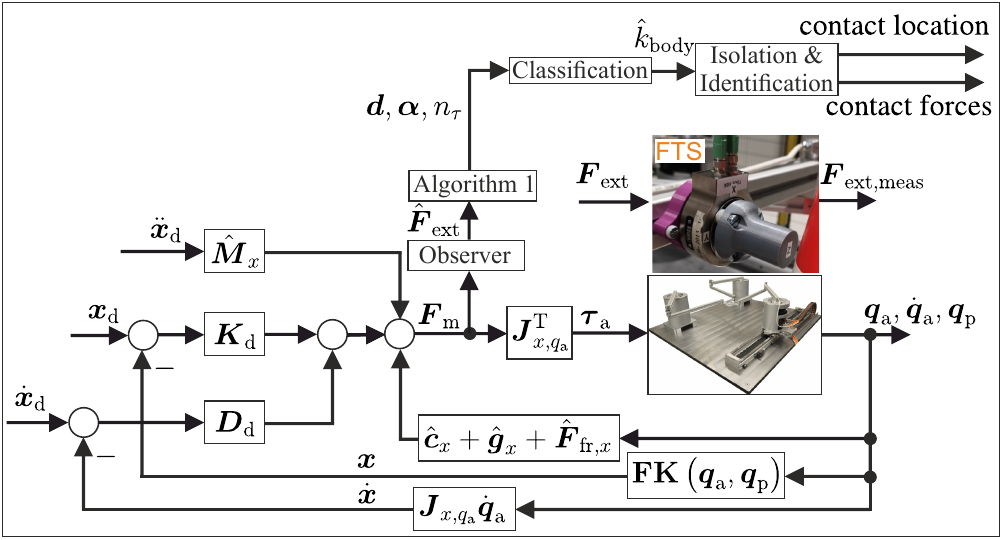}
			\caption{Block diagram with an extended experimental setup from~\cite{Mohammad.2023}}
			\label{fig:Ablauf}
			\vspace{-1.5mm}
		\end{figure}
		Figure~\ref{fig:Ablauf} shows the block diagram of the system which is operated at a sampling rate of $\SI{1}{\kilo\hertz}$. 
		The observer gain of the MO is $k_{\mathrm{o},i}{=}\frac{1}{\SI{50}{\milli \second}}$.
		The stiffness of the Cartesian impedance controller~\cite{Taghirad.2013, Ott.2008} is set to $\boldsymbol{K}_\mathrm{d}{=}\mathrm{diag}(\SI{2}{\newton/ \milli\meter}, \SI{2}{\newton/\milli\meter}, \SI{85}{Nm/\radian})$. 
		Due to the direct drives and the low gear friction, torque control via the motor current is possible. 		
		Critical damping is achieved using the factorization damping design~\cite{AlbuSchaffer.2003}. 
		More information on the test bench is presented in the authors’ previous work~\cite{Mohammad.2023}.
	\subsection{Simulated Results} \label{ssec:SimRes}
		\begin{figure}[bt!]
			\vspace{1.5mm} 
			\centering
			\includegraphics[width=\columnwidth]{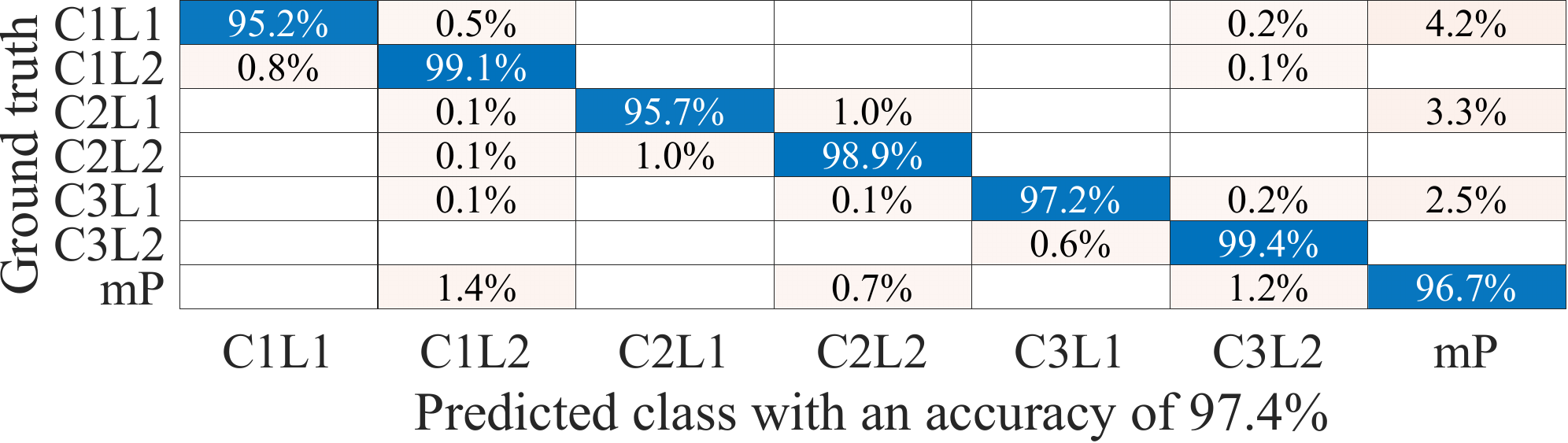}
			\caption{Row-normalized results of the simulation evaluating Alg.~\ref{alg:classifyContBody_DT} (C$i$L$j$ is the $j$-th link of the $i$-th kinematic chain and mP the mobile platform)}
			\label{fig:Simulated_Classification_AffectedBody}
			\vspace{-1.5mm}
		\end{figure}
		In the following, the DT from Alg. \ref{alg:classifyContBody_DT} is investigated in a simulation study. 
		For this, 8154 singularity-free configurations and different $\boldsymbol{F}_\mathrm{ext}$ with $f_{\max{/}\min}{=}{\pm}\SI{140}{\newton},m{=}0$ are simulated. 
		The point of force application and the affected body of the PR are selected randomly and equally distributed from the seven possible classes. 
		By the Jacobian matrices in (\ref{eq:contactJacobian}), $\boldsymbol{F}_\mathrm{ext,link}$ is projected onto the platform and actuated joint coordinates. 
		All inputs for the algorithms \ref{alg:classifyContBody_geomFeat} and \ref{alg:classifyContBody_DT} are defined with the shifted configuration $\boldsymbol{x}{=}\boldsymbol{x}_\mathrm{d}{+}\boldsymbol{K}_\mathrm{d}^{-1} \boldsymbol{F}_\mathrm{ext}$ and corresponding $\boldsymbol{q}$.
		Figure~\ref{fig:Simulated_Classification_AffectedBody} depicts the simulated results of the collision-body classification of stationary and ideal cases with known parameters of the rigid body dynamics. 
		Here, an accuracy of $\SI{97.4}{\percent}$ is achieved, showing the theoretical feasibility of the classification.
		It is noticeable that the errors are higher for the contacts at the first links. 
		This is because these contact forces have a low orthogonal component to the first link, resulting in a low effect on the drives.
	\subsection{Experimental Results} \label{ssec:ExpRes}
		\begin{figure}[b!]
			\vspace{1.5mm} 
			\centering
			\includegraphics[width=\columnwidth]{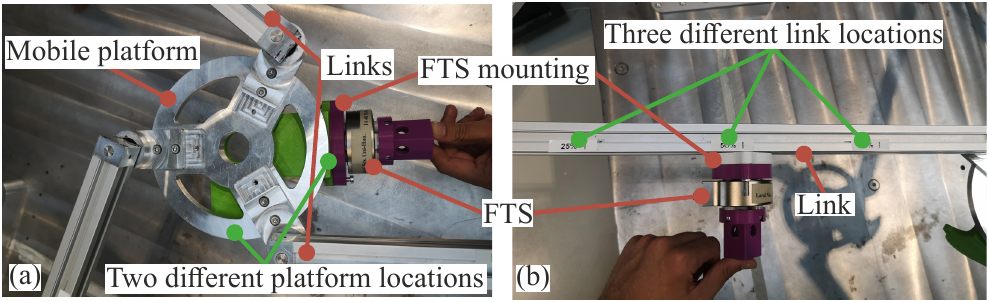}
			\caption{(a) Pushing the FTS mounted on the platform (b) and on a link}
			\label{fig:Montagepositionen}
			\vspace{-1.5mm}
		\end{figure}		
		For the experiments, the FTS is mounted consecutively at three different locations on each of the six links and at two locations on the platform. 
		A push force is applied manually, like in Fig.~\ref{fig:Montagepositionen}. 
		External forces in various directions are then applied to the FTS, causing the PR to respond in an impedance-controlled manner.
		20 sets of measurement data per joint angle configuration are generated for experimental validation of the proposed methods. 
		
		\begin{figure}[t!]
			\vspace{1.5mm} 
			\centering
			\includegraphics[width=\columnwidth]{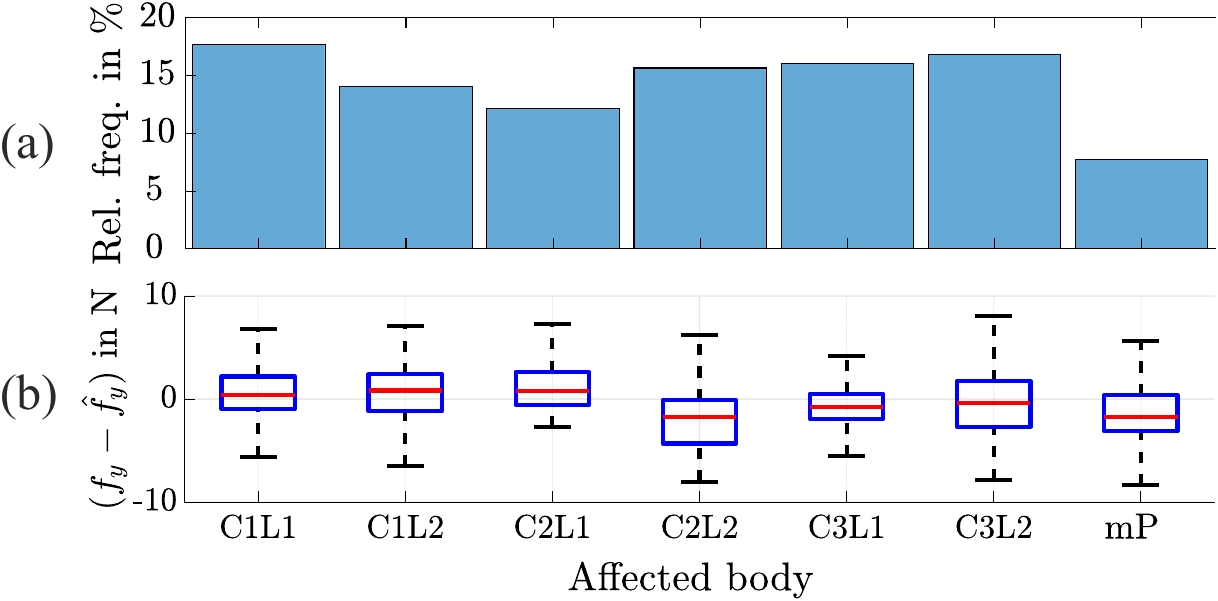}
			\caption{(a) Relative frequency of data points by mobile platform and links. (b) Box plot of the error between estimated force $\hat{f}_y$ by the MO and measured force $f_y$ via FTS over the affected body}
			\label{fig:RelFreq_AffectedBodies_AND_Estimation_over_MeasurementForce_MPX}
			\vspace{-1.5mm}
		\end{figure}		
	
		The relative frequencies of labeled data points in different configurations and the estimation accuracy of the MO are shown in Fig.~\ref{fig:RelFreq_AffectedBodies_AND_Estimation_over_MeasurementForce_MPX}.
		The database consists of 470k data points with load cases from real experiments with the PR, where the estimation of the MO exceeds the contact detection thresholds from~\cite{Mohammad.2023}.
		In the following, algorithm~\ref{alg:classifyContBody_geomFeat} is evaluated based on the FTS~(\ref{sssec:rslt_fts}) and then the more practical MO~(\ref{sssec:rslt_mo}). 
	\subsubsection{Results based on FTS} \label{sssec:rslt_fts}
		\begin{figure}[b!]
			\centering
			\subfloat[
			Relative frequencies over the distance $d$ of the third chain
				\label{fig:RelFreq_CL_mP_distance}
			]
			{
				\includegraphics[width=0.98\columnwidth]{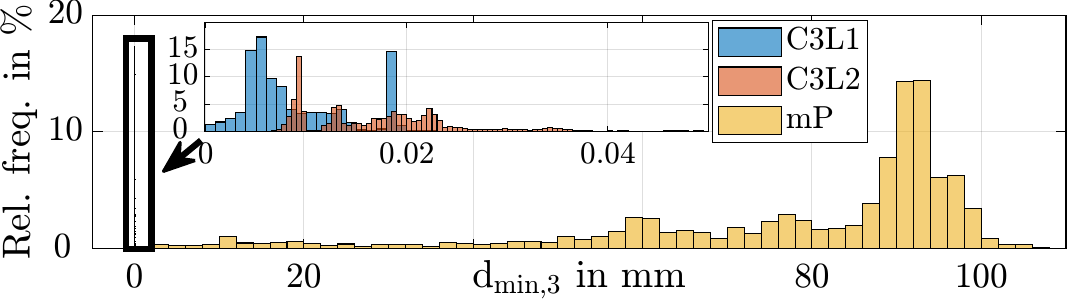} 
			}
			\hspace{-1em}
			\subfloat[
			Relative frequencies over the included angle $\alpha$ of the third chain
			\label{fig:RelFreq_CL_angle}
			]
			{
				\includegraphics[width=0.98\columnwidth]{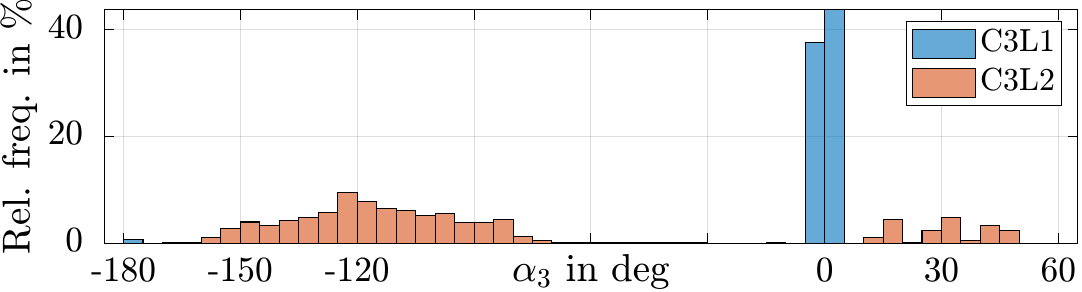} 
			}
			\hspace{-1em}
			\subfloat[
			Color-coded contact cases via $\boldsymbol{\tau}_\mathrm{a,FTS}{=}\boldsymbol{J}_{x_\mathrm{C},q_\mathrm{a}}^\mathrm{T} \boldsymbol{F}_\mathrm{FTS,link}$ 
			\label{fig:AffectedBodies_over_ExtTorques}
			]
			{
			\includegraphics[width=0.98\columnwidth]{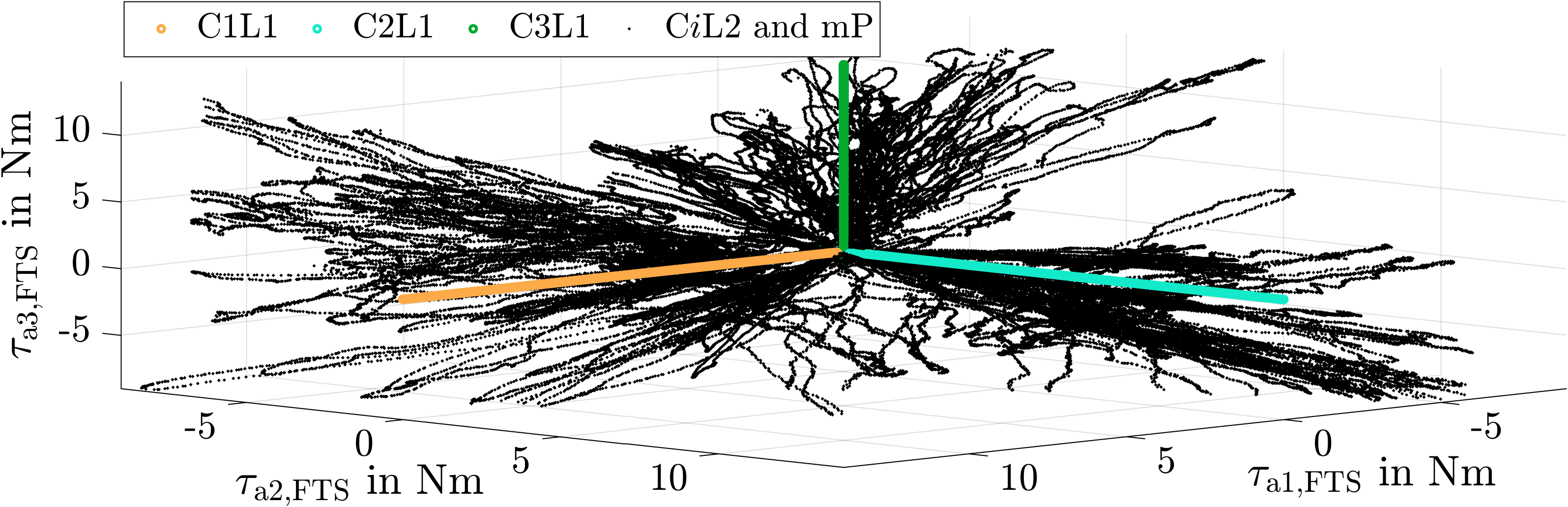} 
			}
			\caption{Results based on $\boldsymbol{F}_{\mathrm{FTS,link}}$ and projections $\boldsymbol{F}_\mathrm{FTS,mP}, \boldsymbol{\tau}_\mathrm{a,FTS}$}
			\label{fig:RelFreq_CL_distance_angle}
		\end{figure}
		In Fig.~\ref{fig:RelFreq_CL_distance_angle}(a)--(b), the relative frequencies of $d_{\min,3}$ and $\alpha_3$ for contacts at the platform, as well as the two links of the third kinematic chain are shown. 
		In the zoom window in Fig.~\ref{fig:RelFreq_CL_distance_angle}(a), a \emph{separation of the platform contacts} $\mathrm{mP}$ (yellow) from the \emph{link contacts} $\mathrm{C3L}i$ (red and blue) is obvious at $d_{\min,3}{=}\SI{0.04}{\milli \meter}$. 
		The underlying reason is the aforementioned effect of the LoA at the coupling joint of the third chain. 
		A \emph{pattern} can be observed in the \emph{link contacts} in Fig.~\ref{fig:RelFreq_CL_distance_angle}(b).
		As described in Sec. \ref{sec:DetAffectedBody}, contacts at the first link are characterized by $\alpha_3{\approx}\SI{0}{\degree}$ (parallel) and $\alpha_3{\approx}\SI{180}{\degree}$ (antiparallel). 
		Another distinction besides $\boldsymbol{d}$ and $\boldsymbol{\alpha}$ is feasible using $\boldsymbol{\tau}_{\mathrm{a,FTS}}$. 
		Figure~\ref{fig:RelFreq_CL_distance_angle}(c) represents $\boldsymbol{\tau}_{\mathrm{a,FTS}}$ in a scatter plot, with contacts at the first links highlighted compared to the rest. 
		This proves that a contact at the first link of the $i$-th chain only acts on the $i$-th drive. 
		
		However, it must be taken into account that the previously determined $\boldsymbol{d}$ and $\boldsymbol{\alpha}$ are based on the projection of the forces measured with the FTS at known locations, which is no condition in a practical scenario. 
		Therefore, the description of the results based on the MO is analyzed in the following.
	\subsubsection{Results based on MO} \label{sssec:rslt_mo}
		\begin{figure}[b!]
			\centering
			\subfloat[
			Relative frequencies over the distance $d$ of the third chain
			\label{fig:RelFreq_CL_mP_distance_BasedOnObserver}
			]
			{
				\includegraphics[width=0.98\columnwidth]{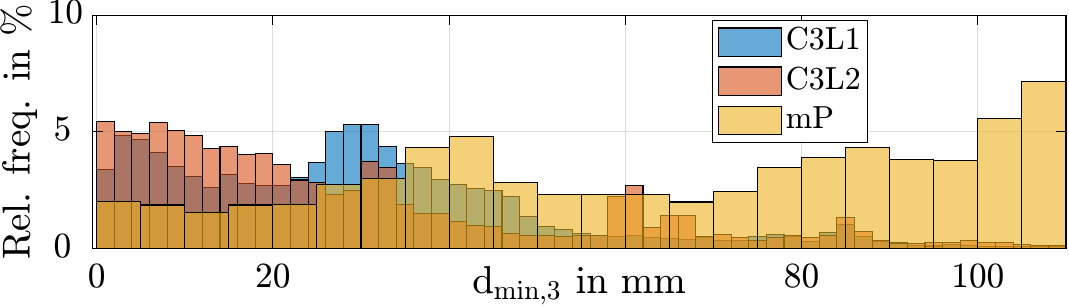} 
			}
			\hspace{-1em}
			\subfloat[
			Relative frequencies over the included angle $\alpha$ of the third chain
			\label{fig:RelFreq_CL_angle_BasedOnObserver}
			]
			{
				\includegraphics[width=0.98\columnwidth]{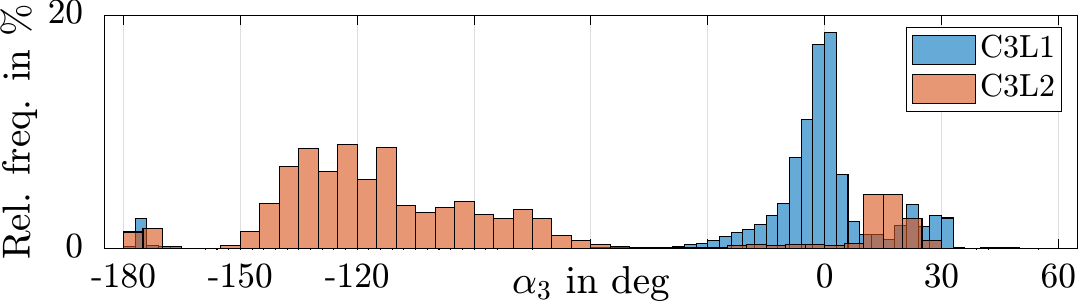} 
			} 
		\hspace{-1em}
		\subfloat[
		Color-coded contact cases via $\hat{\boldsymbol{\tau}}_{\mathrm{a,ext}}{=}\boldsymbol{J}_{x,q_\mathrm{a}}^\mathrm{T}\hat{\boldsymbol{F}}_\mathrm{ext}$ from the MO 
			\label{fig:AffectedBodies_over_ExtTorques_BasedOnObserver}
			]
			{
			\includegraphics[width=0.98\columnwidth]{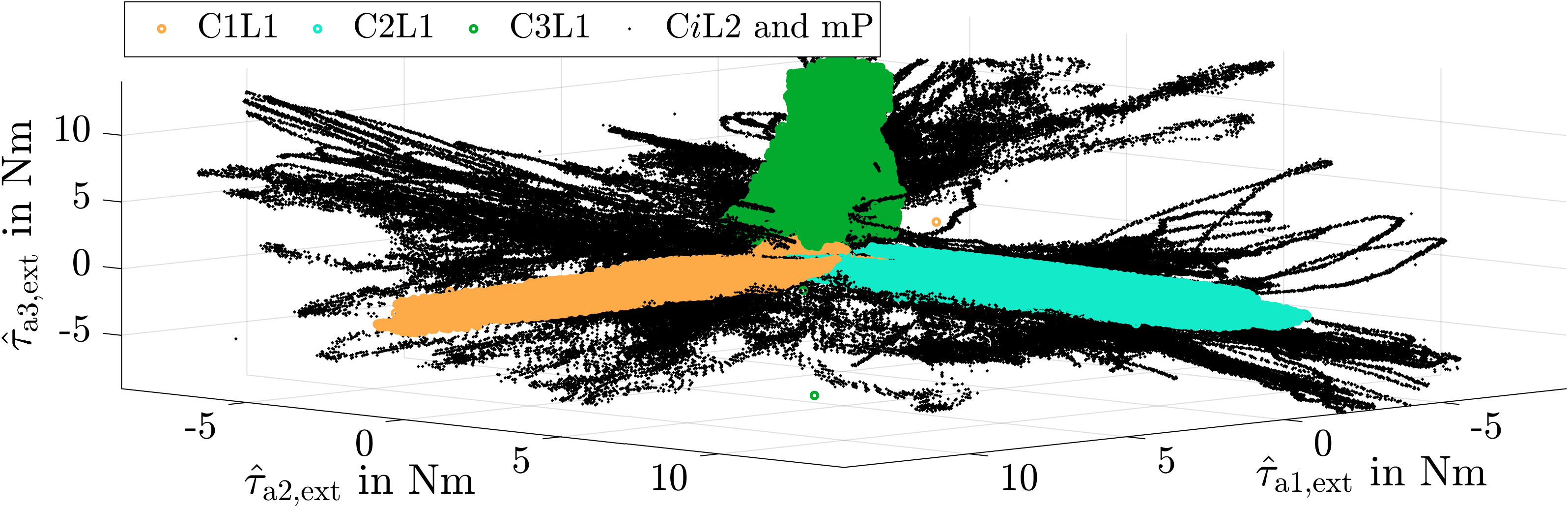} 
			}
			\caption{Results based on the MO}
			\label{fig:RelFreq_CL_distance_angle_BasedOnObserver}
		\end{figure}
		\begin{figure*}[t!]
			\vspace{1.5mm}
			\centering
			\includegraphics[width=\textwidth]{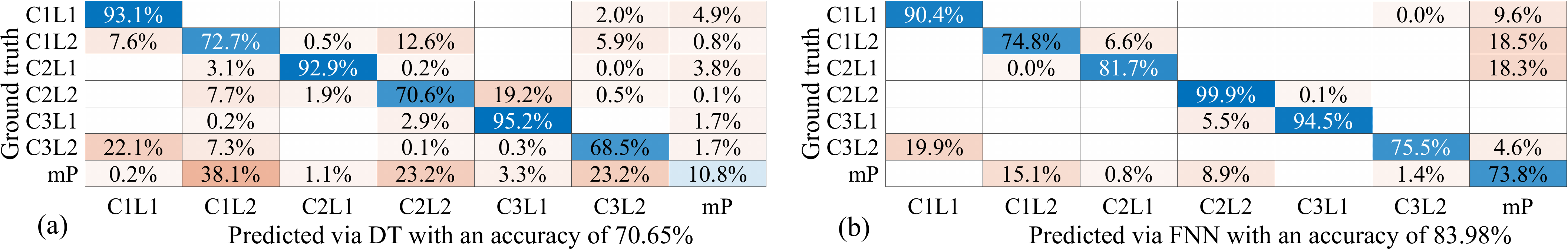} 
			\caption{Row-normalized confusion matrices with test results of collided-body classification (a) with the DT and (b) FNN. Training is performed in only one robot configuration while testing in different ones (C$i$L$j$ is the $j$-th link of the $i$-th kinematic chain and mP the mobile platform)}
			\label{fig:Classification_AffectedBody_Test}
			\vspace{-1.5mm}
		\end{figure*}		
		Figure~\ref{fig:RelFreq_CL_distance_angle_BasedOnObserver} gives the same results representation as Fig.~\ref{fig:RelFreq_CL_distance_angle}, but based on the MO.
		In Fig.~\ref{fig:RelFreq_CL_distance_angle_BasedOnObserver}(a) it can be seen that the platform contact cases have a distance over the entire range of values, while the link contacts are calculated with smaller distances. 
		This may be explained by the inaccurate estimation of the external force (see Fig.~\ref{fig:RelFreq_AffectedBodies_AND_Estimation_over_MeasurementForce_MPX}(b)). 
		Consequently, the position and orientation of the LoA may be distorted to be near a coupling joint.
		In Fig.~\ref{fig:RelFreq_CL_distance_angle_BasedOnObserver}(b), a larger difference occurs at $\alpha_3$ for contacts at the first or second link. 
		Most data points are in the range $\SI{-30}{\degree}{<}\alpha_3{<}\SI{30}{\degree}$ for contacts at C3L1 and in ${\SI{-150}{\degree}{<}\alpha_3{<}\SI{{-}60}{\degree}}$ at C3L2. 
		However, overlaps of both classes are in $\SI{10}{\degree}{<}\alpha_3{<}\SI{30}{\degree}$ and around $\SI{-180}{\degree}$. 
		Particularly in these areas, distinguishing classes based on $\alpha_3$ is ambiguous. 
		
		This can be countered by $\hat{\boldsymbol{\tau}}_\mathrm{a,ext}$, shown in Fig.~\ref{fig:RelFreq_CL_distance_angle_BasedOnObserver}(c).
		Although larger areas for C$i$L1 than the lines in Fig.~\ref{fig:RelFreq_CL_distance_angle}(c) appear, the relationship between the affected chain at the first link and its joint torque is clear. 
		Therefore, $\boldsymbol{\alpha}$ and $\hat{\boldsymbol{\tau}}_\mathrm{a,ext}$ can be combined to classify the ambiguous contact situations as in the range $\SI{10}{\degree}{<}\alpha_3{<}\SI{30}{\degree}$ in Fig.~\ref{fig:RelFreq_CL_distance_angle_BasedOnObserver}(b) more precisely.
		
		As a conclusion to the comparison of the results based on the FTS in Fig.~\ref{fig:RelFreq_CL_distance_angle} and the MO in Fig.~\ref{fig:RelFreq_CL_distance_angle_BasedOnObserver}, it can be stated that the insights regarding $\boldsymbol{d}, \boldsymbol{\alpha}, \hat{\boldsymbol{\tau}}_\mathrm{a,ext}$ from the kinetostatic analysis are confirmed in the experiments with both FTS and the MO.
		However, modeling inaccuracies cause ambiguous contact cases. 
		Examples are cogging torques, friction, or the assumption of equal masses of all chains although the FTS is mounted on one chain. 
		
		The following experiments are based only on the estimation of the MO.
		The generalization of the DT and the FNN considering the modeling inaccuracies is now analyzed (\ref{sssec:ClassificationAffectedBodies}).				
		Finally, collision isolation and identification at a second link using the particle filter are presented (\ref{sssec:pf}).
	\subsubsection{Classification of Collided Bodies} \label{sssec:ClassificationAffectedBodies}
		The training dataset consists of 20 sets of measurements at one configuration to determine the parameters to $\varepsilon_{d}{=}\SI{50}{\milli \meter}$, $\varepsilon_{\alpha}{=}\SI{35}{\degree}$, $\varepsilon_{\hat{\tau}_{\mathrm{a}}}{=}\SI{2.5}{\newton\meter}$ of the DT together with the weights of the FNN. 
		Also, a 5-fold cross-validation is performed on the training dataset to determine the FNN's hyperparameters.
		\begin{figure}[b!]
			\vspace{1.5mm}
			\centering
			\includegraphics[width=0.99\columnwidth]{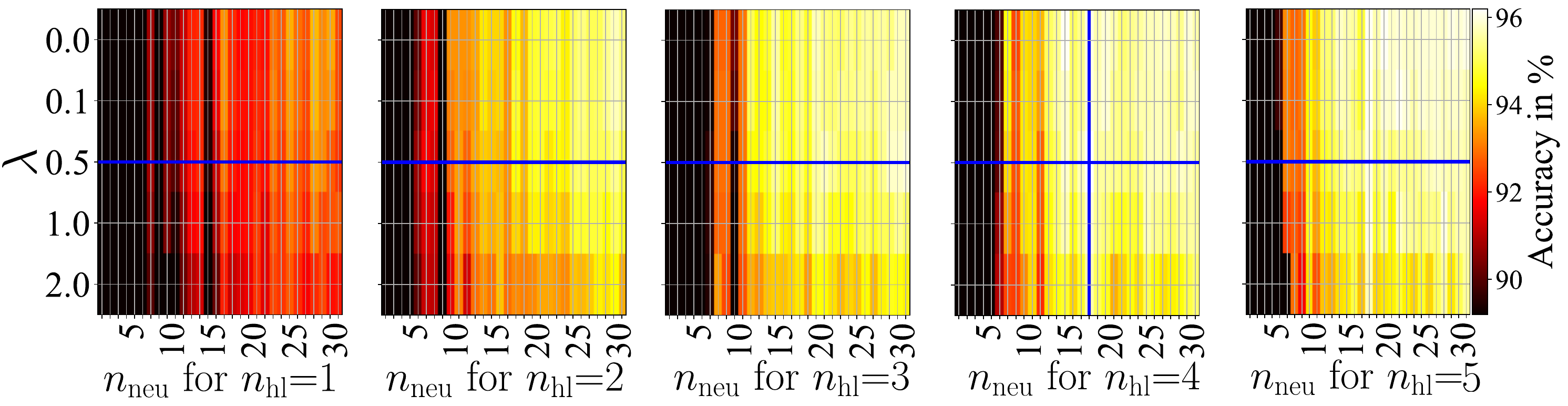}
			\caption{Validation surface of the FNN with blue lines for optimal parameterization}
			\label{fig:ffnn_hypOpt}
			\vspace{-1.5mm}
		\end{figure}
	
		Figure~\ref{fig:ffnn_hypOpt} depicts the cross-validation results for optimizing the regularization factor $\lambda$ and the different numbers of hidden layers $n_\mathrm{hl}$ and neurons $n_\mathrm{neu}$. 
		Up to five hidden layers, each with a maximum of 30 neurons, are compared by classification accuracy. 
		Two tendencies appear from the colored progressions in Fig.~\ref{fig:ffnn_hypOpt}. 
		Network structures with $n_\mathrm{neu}{<}10$ show higher errors, while accuracy improves with increasing $n_\mathrm{hl}$. 
		This is attributed to the increase in the number of weights and nonlinear transformations in the FNN. 
		The intersection of the blue lines corresponds to the selected FNN with $n_\mathrm{hl}{=}4, n_\mathrm{neu}{=}17, \lambda{=}0.5$ and has a collision-body classification accuracy of $96\%$ in the training configuration. 
		
		The generalization of the DT and FNN to two different robot configurations with together 40 sets of measurements is evaluated on the row-normalized test results in Fig.~\ref{fig:Classification_AffectedBody_Test}. 
		In Fig.~\ref{fig:Classification_AffectedBody_Test}(a), the DT performs an accuracy of $71\%$ over the seven bodies including the mobile platform of the PR. 
		However, only $11\%$ of the platform contacts are correctly classified due to the incompletely separable distributions over $\boldsymbol{d}$ (shown in Fig.~\ref{fig:RelFreq_CL_distance_angle_BasedOnObserver}(a)). 
		The reason is that the LoA of the falsely classified platform contacts have a distance less than $\varepsilon_{d}$ from the coupling joints of the kinematic chains. 
		Reducing $\varepsilon_{d}$ would improve the platform classification, but this would lead to a higher chain classification error. 
		
		The FNN has a higher accuracy of $84\%$ compared to the DT in Fig.~\ref{fig:Classification_AffectedBody_Test}(b). 
		In particular, the FNN classifies platform contacts with $74\%$ significantly more accurately. 
		The distinction of the links succeeds most precisely, which is due to the use of $\boldsymbol{\alpha}$ and $\hat{\boldsymbol{\tau}}_{\mathrm{a}}$. 
	\subsubsection{Isolation and Identification} \label{sssec:pf}
		The body classification is followed by the results of the particle filter with $R{=}50$ particles for a contact at a second link. 	
		\begin{figure}[b!]
			\vspace{1.5mm}
			\centering
			\includegraphics[width=1\columnwidth]{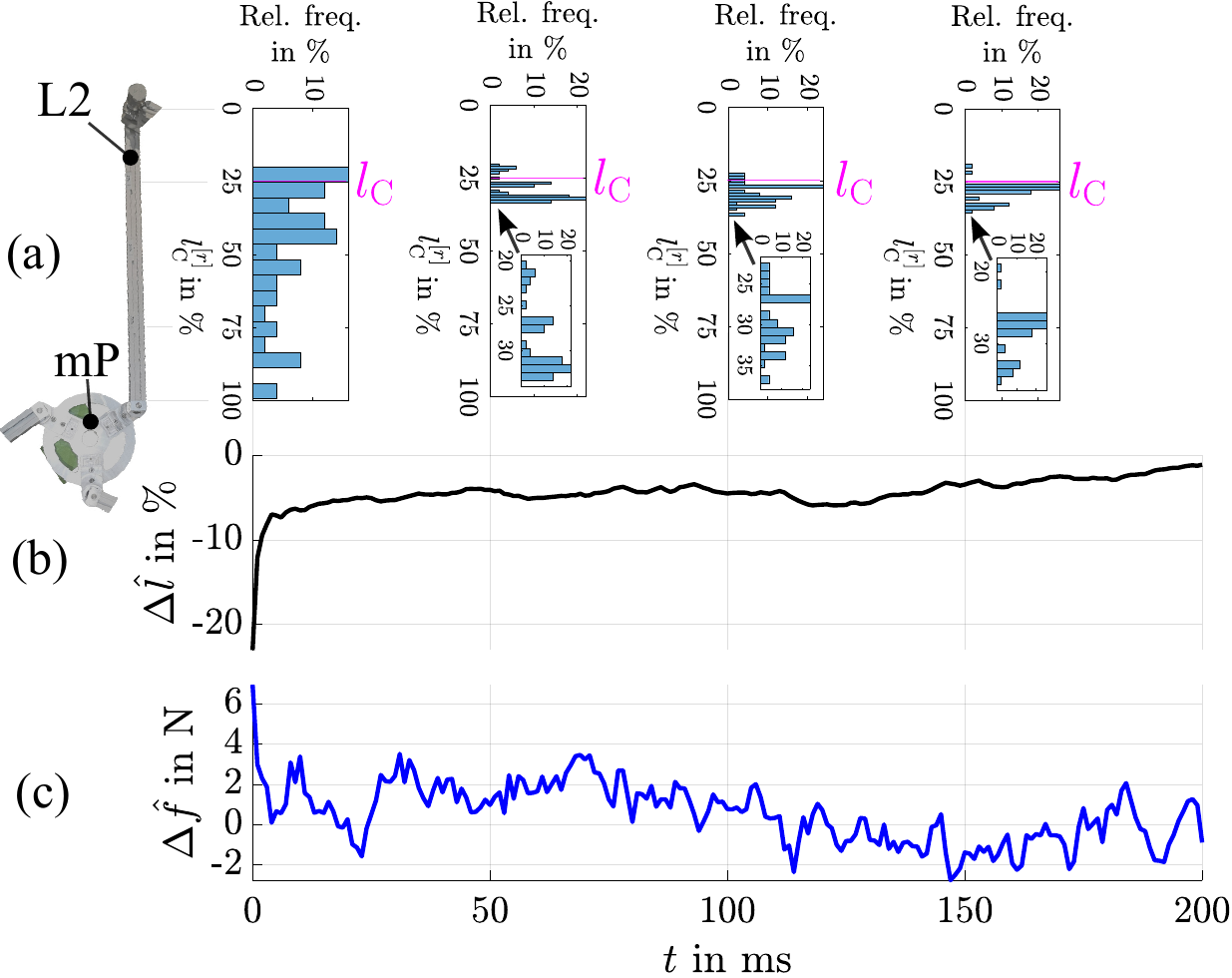}
			\caption{(a) Particle distribution as histograms over the second link at different time steps. (b) Normalized estimation error $\Delta\hat{l}$ and (c) identification error $\Delta \hat{f}$ during a collision at a second link}
			\label{fig:particlefilter}
			\vspace{-1.5mm}
		\end{figure}
		Figure~\ref{fig:particlefilter} shows the time evolution of the particle filter results in an experiment with a push force as shown in Fig.~\ref{fig:Montagepositionen}(b). 
		The trend of the estimation $\hat{l}_\mathrm{C}$ towards the true contact location $l_\mathrm{C}$ (magenta line) is also visible from the time history of the particle distributions in Fig.~\ref{fig:particlefilter}(a).
		In Fig.~\ref{fig:particlefilter}(b), the link-length normalized isolation error $\Delta \hat{l}{=}(l_\mathrm{C}{-} \hat{l}_\mathrm{C})$ is reduced to less than $4\%$ after $\SI{50}{\milli \second}$, corresponding to an error of $\SI{24}{\milli \meter}$. 
		The identification error $\Delta \hat{f}{=}(f_\mathrm{C}{-}\hat{f}_\mathrm{C})$ is below $\SI{4}{\newton}$ after $\SI{50}{\milli \second}$ in Fig.~\ref{fig:particlefilter}(c), which is an order of magnitude less than the defined duration of a transient contact phase~\cite{InternationalOrganizationforStandardization.2016}.
		Thus, the estimation of the contact location and force based on proprioceptive information is successfully applied and could be used in a subsequent reaction.
	\section{Conclusion} \label{sec:conlusions}
		This work aims a proprioceptive collision isolation and identification for parallel robots (PRs). 
		For this purpose, features $\boldsymbol{d}$ and $\boldsymbol{\alpha}$ are derived from the kinetostatic analysis, allowing the classification of a collided body. 
		The simulation results show that the ideal contact body classification by using $\boldsymbol{d}$ and $\boldsymbol{\alpha}$ achieves an accuracy of $97\%$. 
		In real-world experiments, the validity of the features is confirmed for the planar PR. 
		The estimation inaccuracies of the generalized-momentum observer cause ambiguities and thus increase the risk of misclassification.
		With a feedforward neural network, a contact classification accuracy of $84\%$ is achieved in different joint angle configurations based on physically modeled features. 
		Since the links of the PR have a large length-to-radius ratio, the contact isolation for the planar PR is reduced to a one-dimensional problem. 
		Furthermore, a collision force orthogonal to the link is assumed to allow a one-dimensional identification as well. 
		Under these assumptions, a particle filter is developed for collision isolation and identification, which has an error of up to $\SI{3}{\centi \meter}$ and $\SI{4}{\newton}$ after $\SI{50}{\milli \second}$.
		This enables a reaction to the contact location, which will be explored in the future, together with isolation and identification for spatial PRs.
	\addtolength{\textheight}{0cm} 
	


	
	\section*{ACKNOWLEDGMENT}
	The authors acknowledge the support by the German Research Foundation (DFG) under grant number 444769341.
	\bibliographystyle{IEEEtran}
	\bibliography{literatur}	
\end{document}